\newcommand{\method}{{CLaDMoP}}
\newcommand{\methodlong}{{\underline{C}ontrastive \underline{La}nguage \underline{D}rug \underline{Mo}lecule \underline{P}retraining}}
\documentclass[sigconf]
{acmart}

\usepackage{multirow}
\usepackage{algorithm}
\usepackage{algorithmic}
\usepackage{graphicx}
\usepackage{mathrsfs}

\AtBeginDocument{%
  }

\setcopyright{acmlicensed}
\copyrightyear{2025}
\acmYear{2025}
\acmDOI{XXXXXXX.XXXXXXX}
\acmConference[KDD'25]{31th ACM SIGKDD Conference on Knowledge Discovery and Data
Mining}{August 03--07,
  2025}{Toronto, Canada}
\acmISBN{978-1-4503-XXXX-X/2018/06}

\begin{document}

\title{\method{}: Learning Transferrable Models from Successful Clinical Trials via LLMs}

\author{Yiqing Zhang}
\email{yzhang37@wpi.edu}
\affiliation{%
  \institution{Worcester Polytechnic Institute}
  \city{Worcester}
  \state{MA}
  \country{USA}
}

\author{Xiaozhong Liu}
\email{xliu14@wpi.edu}
\affiliation{%
  \institution{Worcester Polytechnic Institute}
  \city{Worcester}
  \state{MA}
  \country{USA}
}

\author{Fabricio Murai}
\email{fmurai@wpi.edu}
\affiliation{%
  \institution{Worcester Polytechnic Institute}
  \city{Worcester}
  \state{MA}
  \country{USA}
}

\renewcommand{\shortauthors}{Y Zhang et al.}

\begin{abstract}
  
Many existing models for clinical trial outcome prediction are optimized using task-specific loss functions on trial phase-specific data. While this scheme may boost prediction for common diseases and drugs, it can hinder learning of generalizable representations, 
leading to more false positives/negatives. To address this limitation, we introduce \method{}, a new pre-training approach for clinical trial outcome prediction, alongside the \underline{S}uccessful \underline{C}linical \underline{T}rials dataset (SCT), specifically designed for this task. \method{} leverages a Large Language Model---to encode trials' eligibility criteria---linked to a lightweight Drug-Molecule branch through a novel multi-level fusion technique. To efficiently fuse long embeddings across levels, we incorporate a grouping block, drastically reducing computational overhead. \method{} avoids reliance on task-specific objectives by pre-training on a ``pair matching'' proxy task. 
Compared to established zero-shot and few-shot baselines, our method significantly improves both PR-AUC and ROC-AUC, especially for phase I and phase II trials.
We further evaluate and perform ablation on \method{} after Parameter-Efficient Fine-Tuning, 
comparing it to state-of-the-art supervised baselines, including MEXA-CTP, on 
the \underline{T}rial \underline{O}utcome \underline{P}rediction (TOP) benchmark. \method{} achieves up to 10.5\% improvement in PR-AUC and 3.6\% in ROC-AUC, while attaining comparable F1 score to MEXA-CTP, highlighting its potential for clinical trial outcome prediction. \textbf{Code and SCT dataset can be downloaded from} {\sloppy\url{https://github.com/murai-lab/CLaDMoP}.}
\end{abstract}




\received{10 February 2025}
\received[revised]{20 May 2025}
\received[accepted]{16 May 2025}

\maketitle

\section{Introduction}

Clinical trial outcome prediction is a challenging task due to the complexity of biological systems. 
Specifically, factors such as heterogeneity~\cite{kravitz2004evidence, dugger2018drug} among patients, unclear disease mechanisms, and differing responses to treatments make accurate predictions particularly difficult. Additionally, regulatory and ethical constraints~\cite{levine1988ethics, edgar2018new}, lengthy approval processes~\cite{chow1998overview}, and the high costs of recruitment further increase time and expenses~\cite{bjornson1993monitoring, dimasi2003price, kakumanu2019cost}.
Despite ongoing challenges faced by the traditional clinical trial prediction methods, the growing availability of historical data on both successful and failed drugs offers a unique opportunity to leverage sophisticated artificial intelligence models for more accurate predictions. 
These advanced methods could improve prediction success rates, enabling more effective resource allocation, prioritizing trials with a higher likelihood of positive outcomes, and ultimately streamlining the drug development process.

\begin{figure}[t]
\centering 
\includegraphics[width=\linewidth]{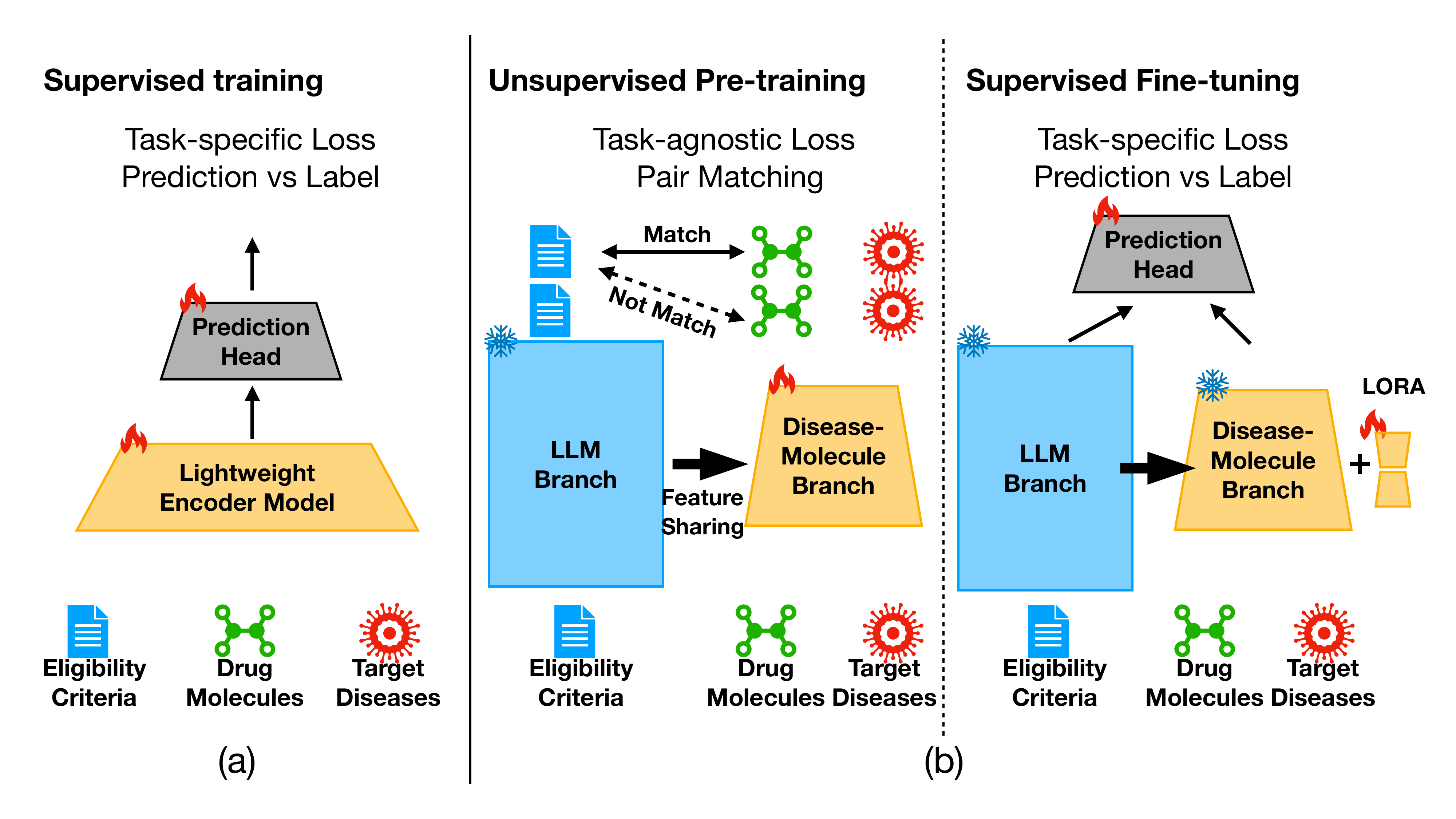}
\Description[<short description>]{<long description>}
\caption{
(\textit{a}) Recent work has focused on using a lightweight encoder model and a prediction head trained with a task-specific loss due to limitations in the availability of labeled data. (\textit{b}) In contrast, \method{} follows a two-stage training paradigm. It incorporates a large language model (LLM branch) and a lightweight attention model (Disease-Molecule branch), sharing knowledge from the former to the latter. (\textit{b-left}) Pre-training stage: uses pair matching as a proxy task and a task-agnostic loss for optimization. (\textit{b-right}) Fine-tuning stage: evaluation performance is further improved by training a prediction head and LoRA layers for DM branch.} 
\label{fig:intro}
\end{figure}

Recent deep learning approaches for clinical trial outcome prediction mainly focus on improving mechanisms for harnessing multi-modality data. 
For example, Hierarchical Interaction Network (HINT)~
\cite{fu2022hint} integrates multimodal data --- such as drug molecules, disease information, trial protocols, and wet lab data (including pharmacokinetics properties like absorption, distribution, metabolism, excretion, and toxicity) --- by extracting and fusing their representations through a human-designed graph model to enhance outcome predictions.
MEXA-CTP~\cite{zhang2025mexa} employs soft and hard masks to selectively extract relevant information. The domain-conditioned masks are optimized by Cauchy loss~\cite{mlotshwa2022cauchy} and leverage masked cross-attention layers to model interactions between pairs of domains, each represented by a module referred to as ``mode expert''. This framework enables the model to learn relationships between drug molecules, diseases, and trial protocols directly from data, rather than relying on hard-coded human priors.

\noindent\textbf{Limitations of State-of-the-Art Approaches.} 
Many models are optimized using task-specific loss functions, which can hinder the learning of robust representations. This is especially problematic for rare or new diseases, for which limited labeled data is available. 
Furthermore, while a task-specific loss function may improve performance for common diseases and drugs, it leads to a decrease in the model’s ability to generalize to special cases (increase in false positives or false negatives), limiting its robustness and adaptability to new, unseen scenarios. Meanwhile, all prior works, like HINT and MEXA-CTP, heavily rely on task-specific losses, which require substantial human effort to manually annotate historical data.


\noindent\textbf{Our Approach}. We present a first-in-class method, \methodlong{} (\method{}), aimed at improving clinical trial outcome prediction performance by incorporating self-supervised learning as a pre-training stage. 
To create the pre-training dataset, we link trials from ClinicalTrials.gov\footnote{\url{https://clinicaltrials.gov/}} with drug synonyms from DrugBank\footnote{\url{https://go.drugbank.com/}}, ensuring all Successful Clinical Trials (SCT) are accurately represented.
Details can be found in Section~\ref{ST}.

Similar to the structure of the CLIP model~\cite{radford2021learning}, \method{} consists of two branches (see Figure~\ref{fig:intro}(b)): the \underline{L}arge \underline{L}anguage \underline{M}odel (LLM) branch utilizes an LLM to extract information from eligibility criteria, while the \underline{D}rug-\underline{M}olecule (DM) branch features a custom transformer-based model designed to learn a joint representation from drug molecules and target diseases.

We freeze the LLM branch and optimize the parameters in the DM branch to maximize the similarity between successful pairs of eligibility criteria embeddings and drug-disease embeddings, while minimizing similarity to all other combinations. We treat the pair matching of all successful clinical trials as a proxy task, and adopt the InfoNCE loss~\cite{oord2018representation} to optimize this objective. 
The LLM branch consists of transformers blocks divided into multiple levels. Every level, except the last, shares distilled knowledge with the DM branch through multi-level fusion. This would typically incur high computational costs due to the increase in token sequence length at every level, causing the computational cost of attention layers to grow disproportionally. To address this issue, we introduce a grouping layer that reduces the sequence length by projecting them to trainable centroid tokens via cross-attention. By stacking multiple grouping layers to form a grouping block, \method{} progressively reduces the number of tokens while maintaining superior performance compared to HINT and MEXA-CTP.


We further enhance the performance of our model via Parameter-Efficient Fine-Tuning (PEFT)~\cite{han2024parameter}. We fix the parameters from the pre-training stage and train the Low-Rank Adaptation (LoRA)~\cite{hu2021lora} layers for the attention layers in DM branch, along with a prediction head. The proposed method outperforms the current state-of-the-art results on the Trial Outcome Prediction (\textsc{TOP}) benchmark.

In sum, \textbf{our main contributions are:}
\begin{itemize}
    \item 
    We propose \method{}, a new pre-training approach for clinical trial outcome prediction, and construct the Successful Clinical Trials (\textsc{SCT}) dataset, specifically designed for the pre-training task.
    Compared to established zero-shot and few-shot baselines, our method demonstrates significant improvements in PR-AUC and ROC-AUC, particularly in phase I and phase II. 
    \item We incorporate a pre-trained LLM model in \method{}, paired with a lightweight model in the DM branch. We extract intermediate LLM embeddings and fuse them to the embeddings from the DM branch through a novel multi-level fusion technique. By implementing a grouping block to aggregate the  information and control the sequence length of intermediate tokens, we drastically reduce the overall training complexity.
    

    \item We evaluate and perform ablation on \method{} after PEFT on the downstream task against several state-of-the-art baselines, including MEXA-CTP, using the \textsc{TOP} dataset from phase I, II, and III clinical trials. \method{} achieves up to 10.5\% and  3.6\% improvement in terms of PR-AUC, and ROC-AUC, respectively, while matching MEXA-CTP's F1 score.
    
    \item We analyze the accuracy and absolute gain in correct predictions for new diseases, comparing \method{} with the previous state-of-the-art MEXA-CTP. We demonstrate that \method{} learns generalizable embeddings through task-agnostic loss and can seamlessly adapt to new clinical trials. Our comparison with MEXA-CTP on the subset of  new diseases reveals that \method{} outperforms MEXA-CTP by 13.63\% in phase I, 8.02\% in phase II, while matching its performance in phase III in terms of accuracy. 
\end{itemize}
\section{Preliminaries}
This section formally defines key elements of the clinical trial outcome prediction task and the notation used throughout the paper.

The standard clinical trial outcome prediction task involves \textbf{Drug Molecules}
$\mathcal{M}$, \textbf{Target Diseases} $\mathcal{D}$ and \textbf{Eligibility Criteria} $\mathcal{C}$. Given a combination of $(\mathcal{M}, \mathcal{D}, \mathcal{C})$, our goal is to predict the final outcome of the clinical trial, i.e.\ whether it succeeded or not.
\begin{definition} \textbf{Drug Molecules} refer to chemical compounds that are designed or discovered to have therapeutic effects on biological systems. They can interact with specific biological targets, such as proteins or enzymes, to modify physiological processes and treat diseases. We denote the set of drug molecules in clinical trial $j$ without considering quantities or percentuals\footnote{In clinical trial data, ingredients' quantities or percentuals are not always informed~\cite{umscheid2011key}, as the main focus is on the presence of molecules relevant to the study.} by 
%
\begin{equation}
    \mathcal{M}^{(j)} := \{M_{1}^{(j)}, M_{2}^{(j)}, \cdots, M_{M_j}^{(j)}\}, \quad j \in [1..N],
\end{equation}
where  each $M_{i}^{(j)}; i \in [1..M_j]$ is a drug molecule which might have an effect on one or more target diseases. 

\end{definition}

\begin{definition} \textbf{Target Diseases} are coded using the ICD-10 (International Classification of Diseases, 10$^\mathrm{\textit{th}}$ revision\footnote{\url{https://www.icd10data.com/ICD10CM/Codes}}) system, which provides a hierarchical structure for categorizing diseases based on their characteristics (e.g., infectious diseases, cardiovascular diseases, cancers). We denote the combination of target diseases in clinical trial $j$ by
\begin{equation}
        \mathcal{D}^{(j)} := \{D_1^{(j)}, D_2^{(j)}, \cdots, D_{d_j}^{(j)}\}
\end{equation}
where $D_i^{(j)}; i \in [1..d_j]$ is an individual ICD-10 diagnosis code.
\end{definition}

\begin{definition} \textbf{Eligibility Criteria} are specific requirements that participants must meet to be included in the study.
They are typically structured text data, including inclusion criteria and exclusion criteria, collectively denoted by $\mathcal{C}^{(j)}$ for clinical trial $j$.
\end{definition}
To avoid clutter, we will omit $(j)$ unless there is risk of ambiguity.

\begin{definition} \textbf{Clinical Trial Outcome.}
Clinical trials typically evaluate the effectiveness of a drug based on a variety of metrics related to the symptoms they aim to alleviate, as well as broader health and safety considerations. These metrics can include specific symptom scores, biomarker levels, or other health indicators. The outcome is often a binary label $y$, where a positive outcome $y=1$ indicates success (i.e., drug was effective or safe), and a negative outcome $y=0$ indicates otherwise. 
\end{definition}


\begin{definition} \textbf{(Clinical Trial) Pre-training}. 
We propose a new task where the goal is to minimize a joint loss $\mathcal{L}_{\mathrm{pre-training}}$ across different modalities, encouraging the model to learn a robust representation. The objective function is defined as:
\begin{equation}
    \theta^*, \phi^*, = \arg \min_{\theta, \phi}\; \mathcal{L}_{\mathrm{pre}-\mathrm{training}}(f(\mathcal{M}, \mathcal{D}; \theta), g(\mathcal{C}; \phi)),
\label{pre-traing}
\end{equation}
where $f(\mathcal{M}, \mathcal{D}; \theta)$ learns a joint representation from drug molecules ($\mathcal{M}$) and target diseases ($\mathcal{D}$), and is parameterized by $\theta$; $g(\mathcal{C}; \phi)$ learns to represent eligibility criteria $\mathcal{C}$, and is parameterized by $\phi$.
In our approach, we fix parameters $\phi$ (LLM branch) during training.
\end{definition}


\begin{definition} \textbf{(Clinical Trial) Fine-tuning.} 
The fine-tuning objective is to minimize a supervised loss function $\mathcal{L}_{\mathrm{fine-tuning}}$ by optimizing the model's parameters. The task is formalized as:
\begin{equation}
    \psi^* = \arg \min_{\psi} \mathcal{L}_{\mathrm{fine}-\mathrm{tuning}} \;(H(f(\mathcal{M}, \mathcal{D}), g(\mathcal{C})); \psi),
\label{fine-tuning}
\end{equation}
where  $H$ is a prediction head parameterized by $\psi$, and the loss function $\mathcal{L}_{\mathrm{fine}-\mathrm{tuning}}$ is designed to optimize the model's performance on the task at hand.
\end{definition}

\section{Proposed Method}
\begin{figure*}[t!] 
\centering 
\includegraphics[width=0.9\linewidth]{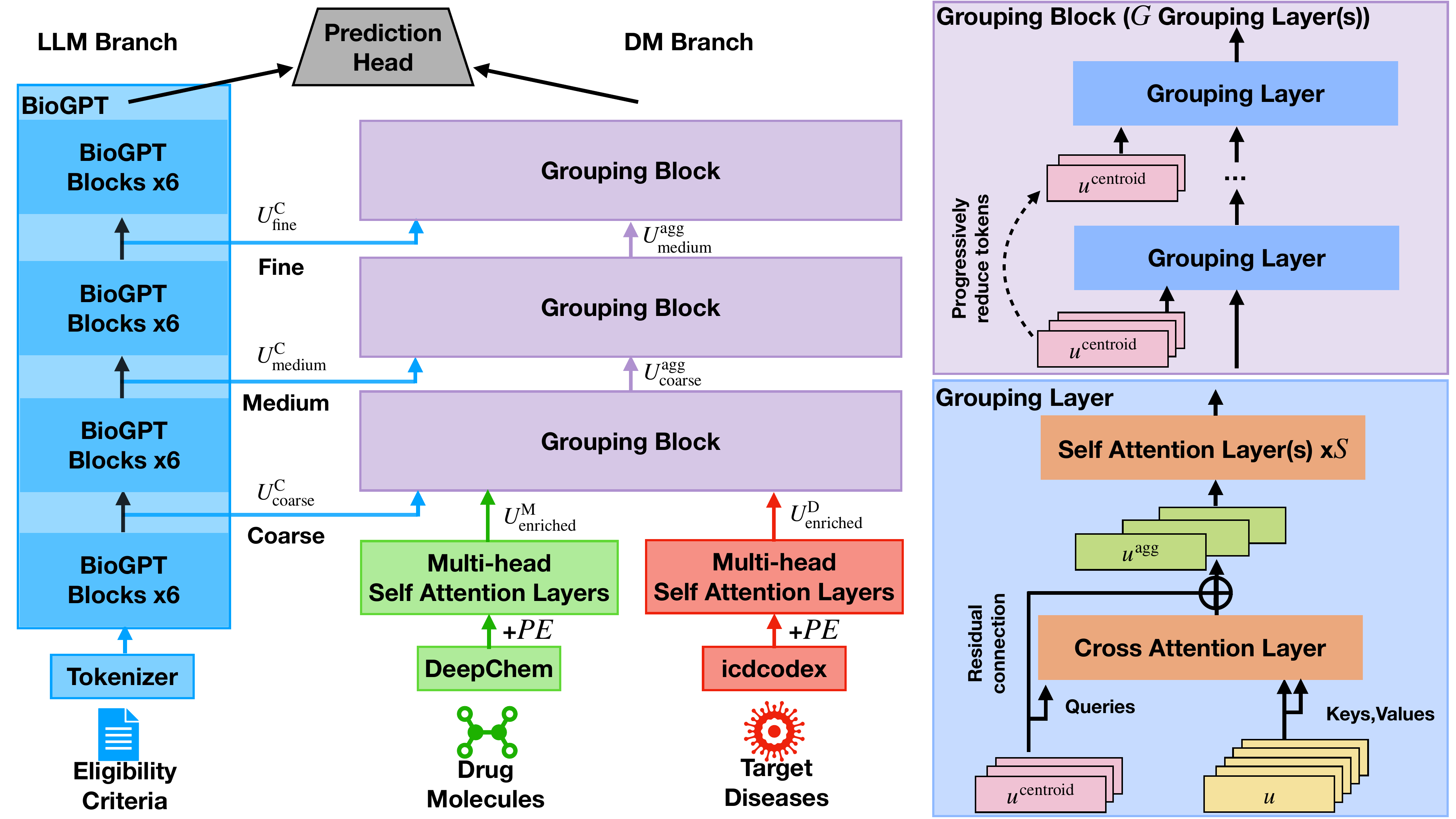}
\caption{(\textit{Left}) \method{} consists of two branches: the LLM branch and the DM branch. The prediction head is a three-layer residual network, which is trained during the PEFT stage.  Further details are provided in Section~\ref{sec:model arc}. (\textit{Right-top}) The grouping block enables the model to fuse information from the LLM branch. Each grouping block consists of 
$G$ grouping layer(s), where the number of centroids is progressively reduced at each layer. The structure of the grouping layer is shown in (\textit{Right-bottom}), while $S$ indicates the number of the self-attention layer in a grouping layer, with further details provided in Section~\ref{sec:grouping}
. Note that $u$ is a single token belonging to token sequence $U$.} 
\label{fig:model}
\end{figure*} 
Inspired by models like CLIP~\cite{radford2021learning} and its variants~\cite{sun2023eva,fan2023improving}, which effectively align vision and language representations, we introduce \method{} to harness the strengths of self-supervised pre-training and large language models. This framework consists of two branches: the \underline{L}arge \underline{L}anguage \underline{M}odel (LLM) branch and the \underline{D}isease-\underline{M}olecular (DM) branch. The LLM branch aims to capture the structural and semantic information from eligibility criteria, while the DM branch focuses on modeling the relationship between drug molecules and target diseases. To fully utilize feature extraction capabilities from LLM model, we design a multi-level feature fusion scheme to integrate embeddings from the LLM to the DM branch. After each level of feature fusion, we apply grouping layers to progressively reduce the number of tokens, preventing linear growth in token length and quadratic growth in attention block complexity during multi-level fusion.

We employ a two-stage training paradigm to enhance clinical trial outcome prediction performance. The first stage involves self-supervised pre-training, where the model learns robust representations for successful pairs of $(\mathcal{M}, \mathcal{D}, \mathcal{C})$ 
via a pair-matching proxy task. In the second stage, we further improve performance with supervised fine-tuning. Our optimization strategy is designed to effectively train the model in under scarce labeled data scenarios.


\subsection{Model Architecture}
\label{sec:model arc}
\subsubsection{LLM branch.}

The LLM branch is responsible for processing  eligibility criteria input. 
We choose BioGPT~\cite{luo2022biogpt} as the LLM model due to its specialization in biomedical text, which enables efficient extraction for the embedding of the eligibility criteria. BioGPT leverages advanced natural language processing (NLP) to understand and interpret domain-specific language, such as inclusion and exclusion criteria, medical history, and other qualification requirements. BioGPT's last layer embedding of the eligibility criteria will be denoted as
$
     U^{\mathcal{C}} = \mathrm{BioGPT}(\mathcal{C})
$.

To further distill knowledge from the BioGPT model, we consider intermediate embeddings obtained at three different levels, going from coarser (shallower) to finer (deeper) representations. Specifically, we consider sequences of 6 consecutive attention blocks, where the first 6 blocks generate coarse level embeddings $U^{\mathcal{C}}_{\mathrm{coarse}}$, then forwarded through the next 6 blocks to generate medium level embeddings $U^{\mathcal{C}}_{\mathrm{medium}}$ and, finally, through the following 6 blocks (for a total of 18 blocks) to generate fine level embeddings $U^{\mathcal{C}}_{\mathrm{fine}}$. 
These multi-level embeddings enable us to capture varying detail and context within the text, enhancing the richness of the representation.

\subsubsection{DM branch}
Learns a joint representation for drug molecules and target disease (details can be found in Appendix~\ref{appd: EM}), aggregating information from the LLM branch.

\noindent\textbf{Drug Molecules Embedding.}
We utilize DeepChem~\cite{ramsundar2018molecular} to encode each molecule expressed as a SMILES string. For different molecules, we add learnable positional embeddings to distinguish whether the segments come from the same molecule. We utilize self-attention layers to further enrich the embedding. The drug molecule embedding is denoted by $U_{\mathrm{enriched}}^{\mathcal{M}}$.

\noindent\textbf{Target Disease Embedding.}
We encode disease hierarchical information using the $\textsc{icdcodex}$ package\footnote{\url{https://github.com/icd-codex/icd-codex}}. We add learnable positional embeddings shared by diseases that belong to the same ICD-10 category. The combined embedding is in turn forwarded through self-attention layers, yielding the final target disease embedding, denoted by $U_{\mathrm{enriched}}^{\mathcal{D}}$. \vspace{1ex}

\noindent\textbf{Multi-level feature fusion.} 
\label{sec:grouping}
To fully incorporate the intermediate embeddings across all levels without significantly increasing the computational cost, we introduce a grouping block to merge and aggregate information from the LLM branch into the DM branch.

\textit{Grouping Layer.} We employ a cross-attention mechanism to integrate information from input tokens into centroids. To facilitate this process, we introduce trainable centroid tokens, initialized from the Gaussian distribution $N(0, 1)$. These centroids generate queries via cross-attention layer, enabling them to efficiently fuse relevant information from the input tokens (i.e., drug molecule, target disease, and eligibility criteria tokens) as follows:
\begin{equation}
U^\mathrm{agg} = \mathrm{softmax}\left(\frac{U^{\mathrm{centroid}}W^Q (UW^K)^\top}{\sqrt{d_k}}\right) UW^V,
\end{equation}
where $U$ represents the input tokens which generate keys and values via $W^K$ and $W^V$, where $U \in \{U_{\mathrm{enriched}}^{\mathcal{M}}, U_{\mathrm{enriched}}^{\mathcal{D}}, {U_{x}^{\mathcal{C}}}, {U_{x-1}^{agg}}\}$, $x \in \{\mathrm{coarse}, \mathrm{medium}, \mathrm{fine}\}$, $U^{\mathrm{centroid}}$ indicates the centroid tokens which generate queries, and $d_k$ is the embedding dimension. Information is aggregated based on the attention map produced by the softmax function. By using this cross-attention mechanism, we ensure that the fused output $U^\mathrm{agg}$ always has the same sequence length as $U^{\mathrm{centroid}}$, which dramatically reduces the computational overhead (a detailed study of the complexity
analysis is presented in Section~\ref{sec:complexity}). 
Additionally, to mitigate potential losses in embedding quality, we stack $S$ self-attention layers on the top of the cross attention layer. This design not only helps manage sequence length but also prevents the loss of critical information. Additionally, to ensure that the model captures input tokens from diverse sources, we add learnable positional embeddings to distinguish them.

\textit{Grouping Block.}
We stack 
$G$ grouping layers to form a grouping block. The purpose of stacking these grouping layers is to progressively reduce the number of centroids. Specifically, in each layer, we reduce the number of centroids by half compared to the previous layer, ensuring that all key information is retained.


The DM branch first fuses $U_{\mathrm{coarse}}^{\mathcal{C}}$, followed by $U_{\mathrm{medium}}^{\mathcal{C}}$ and $U_{\mathrm{fine}}^{\mathcal{C}}$. 
We add layer normalization to the final output to ensure it matches the scale of the LLM branch. The final output tokens of the DM branch are denoted by $U^{\mathcal{D,M}}$.

\subsection{Pre-training Stage}
In the pre-training stage, our goal is to train the model to correctly identify a match when $(\mathcal{M}, \mathcal{D}, \mathcal{C})$ comes from a successful trial. This is equivalent to ensuring that the match has a higher similarity score than both mismatched pairs and failed trials. To reduce the effort required for manual labeling, we adopt a more efficient approach for collecting a new dataset that focuses solely on \underline{S}uccessful \underline{C}linical \underline{T}rials, \textsc{SCT} dataset.

\begin{table}[htbp]
\caption{\underline{S}uccessful \underline{C}linical \underline{T}rials, \textsc{SCT} dataset statistics.}
\label{statistics_sct}
\centering
\setlength\tabcolsep{15pt}
\begin{tabular}{lr}
\toprule
Statistic & Value \\
\midrule
Number of drugs & 4,289 \\
Number of diseases & 3,326 \\
Avg. words in eligibility criteria & 364\\
Unique drug combinations & 2,902 \\
Unique disease combinations & 951 \\
\bottomrule
\end{tabular}
\end{table}

\noindent\textbf{Dataset.} \label{ST}
%
We construct this dataset by linking data from ClinicalTrials.gov\footnote{\url{https://clinicaltrials.gov/}} and DrugBank\footnote{\url{https://go.drugbank.com/}}. While each clinical trial explicitly states the drug name, drug synonyms may appear in related or subsequent trials. To ensure that drug synonyms across multiple trials—especially those related to the same disease—are correctly mapped to the same drug entity, we reference DrugBank to identify synonyms and the target disease. This allows us to establish a sequence of trials for each drug.
For drugs already on the market, we assume that all clinical trials from phase I to phase III were successful~\cite{downing2014clinical, bobo2016nanoparticle}. Additionally, we assume that a drug progressing to phase 
$p \in \{II,III,IV\}$ has successfully completed all previous phases\footnote{\url{https://www.fda.gov/patients/learn-about-drug-and-device-approvals/drug-development-process}}. To maintain data integrity, we automatically filter out trials where drugs lack relevant experiments in subsequent phases. For example, if a drug has results from phase I to phase III but no phase IV data, we exclude it from the phase III dataset, as the outcome remains uncertain without validation of the clinical trial's public records. 
In contrast, our process is fully automated, reducing the need for manual labeling. 
Table~\ref{statistics_sct} shows key statistics of the SCT dataset. To prevent data leakage, we ensure that none of the trials in TOP’s test set are included in SCT’s pre-training data. Specifically, we perform a temporal split based on NCTid for train-test separation, analogous to the split done by previous works for TOP.

We freeze the parameters in the LLM branch and train the DM branch using a pair-matching task, where the goal is to predict the correct pairings  ($U^{\mathcal{C}}$ and $U^{\mathcal{D,M}}$).
To achieve this, we optimize a symmetric cross-entropy loss over the similarity scores, which is widely used in contrastive representation learning as the InfoNCE loss. We adapt this approach specifically for the clinical trial outcome prediction task shown in Algorithm~\ref{algo:pre-training}.
\begin{algorithm}[htbp]
    \caption{Pseudocode for Pre-training Loss}
    \begin{algorithmic}[1]
        \STATE \textbf{Input:} Batch size $n$, temperature parameter $\tau$
        \STATE \textbf{Output:} Pre-training loss $\mathrm{L}_{\mathrm{pre}-\mathrm{training}}$

        \STATE \COMMENT{Identity matrix indicating which pairs are CORRECT matches}
        $\mathrm{labels} = \mathrm{diag}([1, 1, 1, ..., 1])_{n \times n}$
        
        \STATE $f_\mathcal{C}  = \mathrm{Avgpool}(U^{\mathcal{C}})$
        \COMMENT {final eligibility criteria embeddings}

        \STATE $f_\mathcal{DM}  = \mathrm{Avgpool}(U^{\mathcal{D,M}})$
        \COMMENT {final disease-drug embeddings}

        \STATE \COMMENT{Compute logits w/ temperature scaling for all embedding pairs}\\
        $\mathrm{logits} = \exp(\tau) \cdot f_\mathcal{C} \times f_\mathcal{DM}^\top$\\

        \STATE \COMMENT{Compute cross-entropy over rows}\\
        $\mathcal{L}_\mathrm{row} = \mathrm{cross\_entropy}(\mathrm{logits}, \mathrm{labels}, \mathrm{axis}=0)$\\

        \STATE \COMMENT{Compute cross-entropy over columns}\\
        $\mathcal{L}_\mathrm{col} = \mathrm{cross\_entropy}(\mathrm{logits}, \mathrm{labels}, \mathrm{axis}=1)$\\ 
        
        \STATE $\mathcal{L}_{\mathrm{pre}-\mathrm{training}} = (\mathcal{L}_\mathrm{row}+\mathcal{L}_\mathrm{col})/2$ \COMMENT{final pre-training loss}
    \end{algorithmic}
    \label{algo:pre-training}
\end{algorithm}

\subsection{Fine-tuning Stage (using PEFT)}

The goal of the PEFT stage is to enable the model to discriminate between successful and failed trials under supervision. To achieve this, we apply average pooling to both LLM and DM branches in order to aggregate the information in each.
The resulting embeddings from each branch are then concatenated and passed through a prediction head with a sigmoid activation function. This produces a probability score indicating whether the trial is successful or not:
\begin{equation}
    \hat{y} = \sigma\left(H_{\mathrm{pred}}\left(\mathrm{concat}\left(\mathrm{Avgpool}(U^{\mathcal{C}}), \mathrm{Avgpool}(U^{\mathcal{D,M}})\right)\right)\right),
\end{equation}
where $H_{\mathrm{pred}}$ represents the prediction head. To further improve fine-tuning performance without affecting the parameters from the pre-training stage, we freeze the pre-trained parameters and introduce LoRA layers to the DM branch.

A class-weighted binary cross-entropy loss is used for guiding the model training due to the imbalance of negative/positive pairs:
\begin{equation}
    \mathcal{L}_{\mathrm{PEFT}} = -\omega_{0} y\log\hat{y} - \omega_{1} (1-y)\log(1-\hat{y}),
\end{equation}
where $\omega_{0}$ ($\omega_{1}$) is the fraction of negative (positive) labels in the training set.
\label{sec:method}

\section{Experiments}



\begin{figure}[tbp]
\centering 
\includegraphics[width=\linewidth]{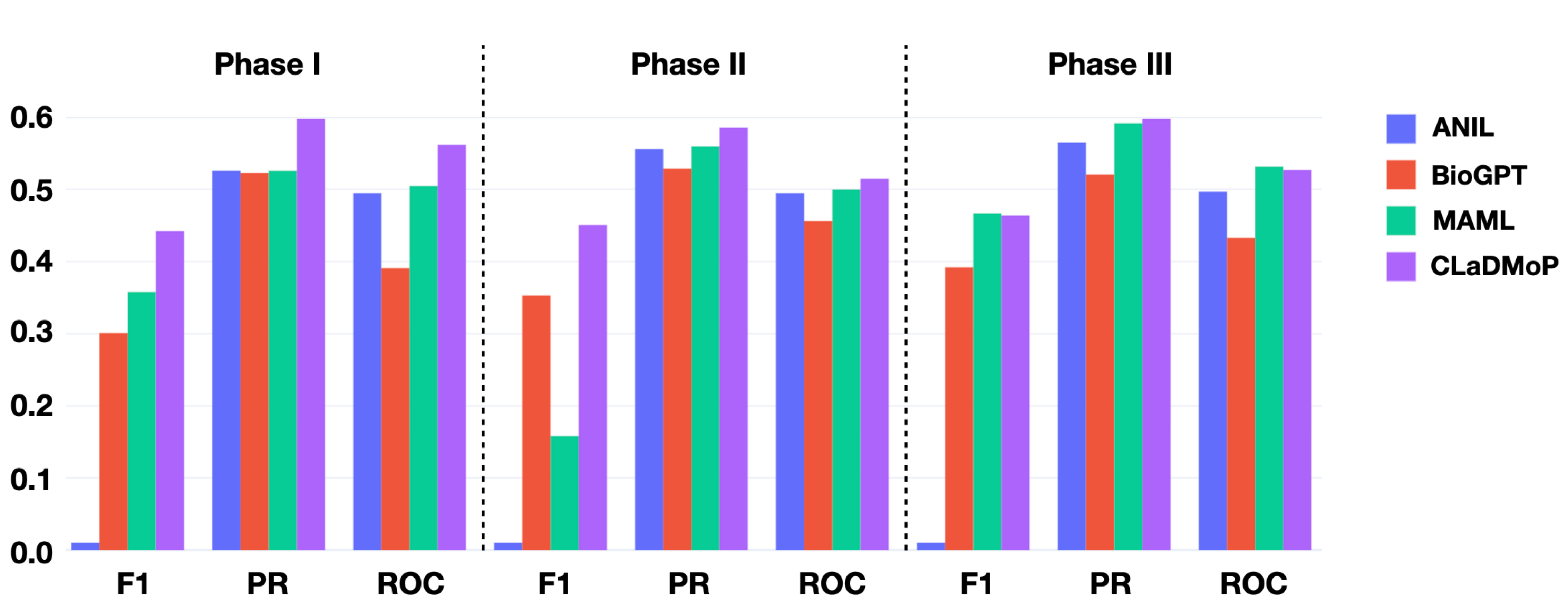}
\caption{Few-shot learning results for clinical trial outcome prediction for phase I, II and III trials.} 
\label{few-shot}
\end{figure}
\begin{table*}[t!]
\caption{Experimental results for outcome prediction for phase I, II, and III trials. Results correspond to averages and standard deviations over 10 bootstrap samples.}
\centering
\setlength\tabcolsep{5pt}
\begin{tabular}{l|lccccccccc}
\toprule
&Metric & LR & RF & KNN+RF & XGBoost & AdaBoost & FFNN & HINT & MEXA-CTP & \method{} \\
\midrule
\parbox[t]{2mm}{\multirow{3}{*}{\rotatebox[origin=c]{90}{Phase I}}} & 
F1 & .503$\pm$.017 & .519$\pm$.017 & .522$\pm$.014 & .624$\pm$.016 & .633$\pm$.015 & .614$\pm$.009 & .604$\pm$.009 & \textbf{.713}$\pm$.027 & \textbf{.713}$\pm$.019 \\
& PR-AUC & .515$\pm$.015 & .514$\pm$.015 & .514$\pm$.004 & .594$\pm$.015 & .544$\pm$.011 & .569$\pm$.012 & .581$\pm$.021 & .605$\pm$.014 & \textbf{.680}$\pm$.016 \\
& ROC-AUC & .514$\pm$.018 & .542$\pm$.014 & .528$\pm$.009 & .538$\pm$.016 & .540$\pm$.012 & .550$\pm$.010 & .575$\pm$.021 & .593$\pm$.012 & \textbf{.642}$\pm$.015 \\
\midrule
\parbox[t]{2mm}{\multirow{3}{*}{\rotatebox[origin=c]{90}{Phase II}}} & 
F1 & .533$\pm$.017 & .534$\pm$.013 & .594$\pm$.011 & .572$\pm$.013 & .579$\pm$.008 & .579$\pm$.008 & .635$\pm$.010 & \textbf{.695}$\pm$.008 & .685$\pm$.015 \\
& PR-AUC & .560$\pm$.012 & .573$\pm$.015 & .583$\pm$.014 & .585$\pm$.015 & .586$\pm$.013 & .586$\pm$.013 & .608$\pm$.011 & .635$\pm$.015 & \textbf{.683}$\pm$.012 \\
& ROC-AUC & .567$\pm$.016 & .576$\pm$.010 & .583$\pm$.016 & .601$\pm$.003 & .589$\pm$.013 & .601$\pm$.012 & .623$\pm$.012 & .638$\pm$.005 & \textbf{.647}$\pm$.018 \\
\midrule
\parbox[t]{2mm}{\multirow{3}{*}{\rotatebox[origin=c]{90}{Phase III}}} & 
F1 & .624$\pm$.013 & .675$\pm$.018 & .670$\pm$.018 & .694$\pm$.017 & .722$\pm$.014 & .625$\pm$.017 & .814$\pm$.013 & .857$\pm$.007 & \textbf{.861}$\pm$.014 \\
& PR-AUC & .553$\pm$.011 & .583$\pm$.024 & .587$\pm$.016 & .627$\pm$.009 & .589$\pm$.015 & .572$\pm$.020 & .603$\pm$.014 & .771$\pm$.016 & \textbf{.860}$\pm$.011 \\
& ROC-AUC & .600$\pm$.028 & .643$\pm$.023 & .643$\pm$.024 & .668$\pm$.014 & .624$\pm$.013 & .620$\pm$.023 & .685$\pm$.023 & .693$\pm$.025 & \textbf{.702}$\pm$.018 \\
\bottomrule
\end{tabular}
\label{results}
\end{table*}

We meticulously follow the experimental settings described in the Trial Outcome Prediction (\textsc{TOP}) benchmark~\cite{fu2022hint} to evaluate our model for both pre-training stage and fine-tuning stage. In addition, to assess the capabilities of the learning components, we conduct ablation studies to evaluate the importance of different components of our model.


\subsection{Experimental Settings}\label{sec:settings} To the best of our knowledge, the \textsc{TOP} benchmark is the only publicly available dataset for clinical trial outcome prediction and has been incorporated into TrialBench for the clinical trial approval task. To evaluate the proposed model, we use the \textsc{TOP} benchmark and follow the same strategy for splitting the train, and test sets. Specifically, for each phase, we split the dataset based on the start day of the clinical trials, allocating earlier trials to the train set and using later trials for model evaluation. 
This approach allows us to evaluate the model's performance on unseen data from different time periods, helping to assess its generalization ability. 
During training, we randomly select $15\%$ of the training samples as the validation set to monitor the model's performance and tune hyperparameters. This protocol is part of the TOP's experimental design, ensuring consistency and fair comparisons with other studies. For statistics of the data splits, please refer to Appendix~\ref{appd: top}.

\subsubsection*{Evaluation Metrics.} \label{eval}We utilize standard evaluation metrics for clinical trial outcome prediction, including F1 score (i.e., the harmonic mean of precision and recall), PR-AUC (i.e., area under the precision-recall curve), and ROC-AUC (i.e., area under the receiver operating characteristic curve). For all three metrics, higher values indicate better performance.

\subsubsection*{Pre-training Stage.} \noindent\textit{Our Model.} We use BioGPT as the eligibility criteria encoder within the LLM branch. In DM branch, we stack four self-attention layers in each attention block to extract features, separately for drug molecules and target diseases. Each attention layer contains 8 attention heads, with an embedding size of 16. We then use 3 grouping blocks to integrate and aggregate information from the LLM branch into the DM branch. Each grouping block consists of a cross-attention layer followed by two self-attention layers. The number of centroids starts at 100 and is gradually halved, ultimately reducing to 25 at the final block.

\textit{Training.} All the parameters in the LLM branch are fixed during pre-training. We optimized the DM branch with Adam using default values for $\beta_1$ and $\beta_2$. Based on our results from hyperparameter tuning (see Appendix~\ref{pre-bs}), we set the batch size to 128 and use a fixed learning rate $10^{-4}$ for training. The temperature for the InfoNCE loss is set to 0.6. Our model was pre-trained on \textsc{SCT} dataset.

\textit{Baselines.} 
We consider several baselines for the pre-training task to evaluate whether the pair-matching task serves as a suitable proxy for the classification problem. Since the optimization problem is an unsupervised learning task, we compare it with both zero-shot (ANIL~\cite{raghu2019rapid} and pre-trained BioGPT~\cite{luo2022biogpt}) and few-shot (MAML~\cite{finn2017model}) learning approaches. To train these meta-learning models, we further split the \textsc{TOP} benchmark into different meta-tasks based on chronological order. 
For each meta-task, there are 100 clinical trials, from which we randomly select 10 for the support set and the remaining 90 for the query set. We train ANIL (zero-shot) and MAML (few-shot) models as baselines using this split. Additionally, we consider the pre-trained BioGPT (zero-shot) as another baseline. 
Description and implementation details can be found in Appendix~\ref{app: pre-training baselines}.

\textit{Evaluation.} All results from the pre-training stage are evaluated on the query set of the \textsc{TOP} benchmark. MAML is the only method that uses the support set during testing. 

\subsubsection*{PEFT Stage.} \textit{Our model.} We add a prediction head, consisting of a 3-layer feed-forward neural network with ReLU activation and residual connections. LoRA layers with a rank size of 8 are added to all attention layers in the DM branch. The best model parameters from pre-training stage are used as the initial weights, and we continue training only the prediction head and LoRA layers.

\textit{Optimization.} We use the Adam optimizer with default values for $\beta_1$ and $\beta_2$. The batch size is set to 128, with a fixed learning rate of $10^{-2}$ for the prediction head, and $5x10^{-2}$ for the LoRA layers during phase I and phase II, $10^{-3}$ for the phase III training.

\textit{Baselines.} We compare \method{} with several baselines often evaluated on the TOP benchmark, including Logistic Regression (\textbf{LR}), Random Forest (\textbf{RF}), k-Nearest Neighbor + Random Forest (\textbf{kNN+RF}), \textbf{XGBoost}, Adaptive Boosting (\textbf{AdaBoost}), a 3-layer Feed-Forward Neural Network (\textbf{FFNN}), \textbf{HINT}~\cite{fu2022hint} and \textbf{MEXA-CTP}~\cite{zhang2025mexa}. Detailed information can be found in the Appendix~\ref{app: fine-tuning baselines}.

\textit{Evaluation.} We use bootstrapping to evaluate
the model performance 10 times on a random selection
of 80\% of the test data to report the mean and standard
deviation of the evaluation metrics.

\subsection{Experiment
Results}\label{sec:results}
\noindent\textbf{Pre-training Results.} Figure~\ref{few-shot}
shows the comparison results for each phase. The performance gaps over the best-performing baseline (MAML) are larger in the first two phases: 23.5\% and 184.8\% for F1; 13.7\% and 4.6\% for PR-AUC; and 11.3\% and 3.0\% for ROC-AUC. 
The performance gaps over MAML are smaller in phase III, as the dataset grows larger, allowing MAML to learn more effectively from the support task. From the training perspective, our model is more stable than MAML and ANIL across all phases training. As shown later, our pre-trained model outperforms even some supervised models, demonstrating the effectiveness of our approach.

\noindent\textbf{Fine-tuning Results.} Table~\ref{results} shows the comparison results for phase I, phase II, and phase III including standard deviation values obtained via bootstrapping. For all phases, \method{} consistently achieves the best performance in terms of PR-AUC and ROC-AUC, immediately followed by MEXA-CTP.
For comparison purposes, we compute simple (unweighted) averages of their performance differences across phases. On average, \method{} achieves 10.5\% and 3.6\% higher PR-AUC and ROC-AUC, while maintaining a comparable F1 score
to that of MEXA-CTP.


\begin{figure}[t]
\centering 
\includegraphics[width=\linewidth]{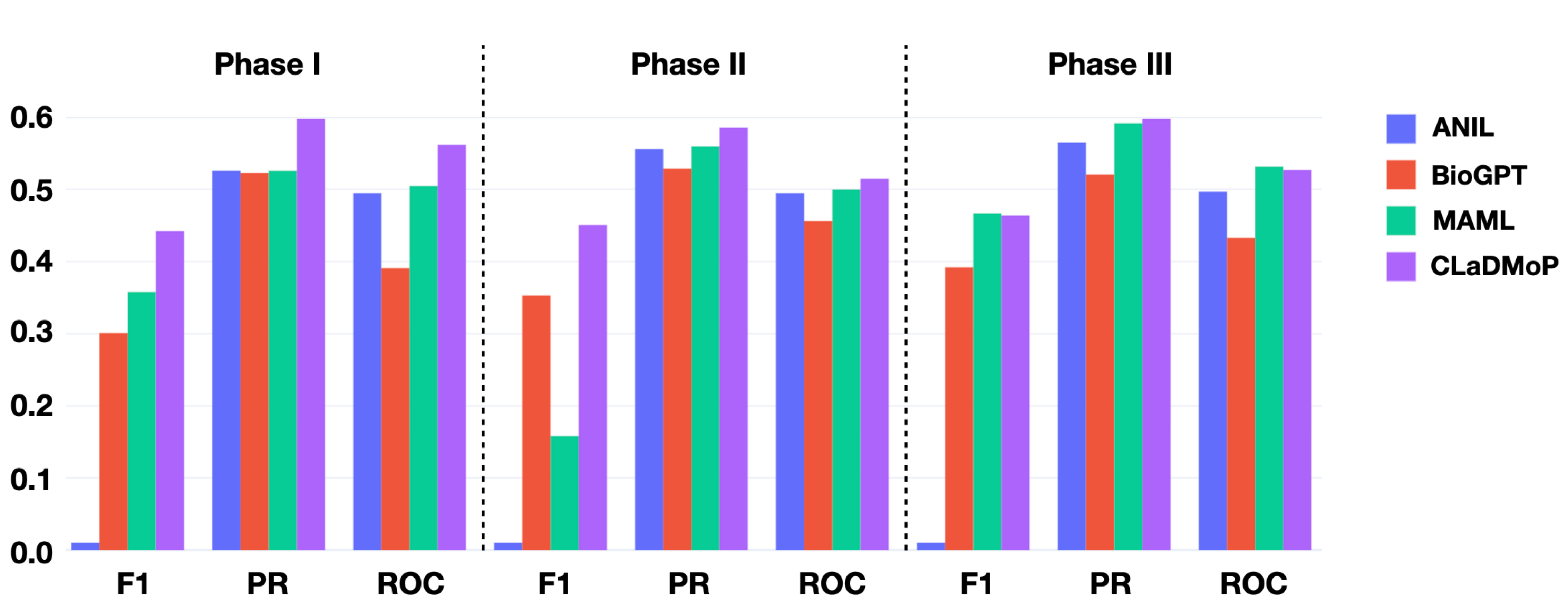}
\caption{Few-shot learning results for clinical trial outcome prediction for phase I, II and III trials.} 
\label{few-shot}
\end{figure}

\begin{table}[t!]
\caption{Evaluation results comparing MEXA-CTP and \method{} on new diseases. Best results are highlighted in bold.}
\centering
\setlength\tabcolsep{5pt}
\begin{tabular}{l|lccccc}
\toprule
&Method & F1 & PR-AUC & ROC-AUC & ACC & Gain \\
\midrule
\parbox[t]{2mm}{\multirow{2}{*}{\scriptsize \rotatebox[origin=c]{90}{Phase I}}} 
& MECA-CTP & .428 & .522 & .517 & 47.55 & \multirow{2}{*}{+9}\\
& \method{} & \textbf{.448} & \textbf{.535} & \textbf{.520} & \textbf{54.34} & \\
\midrule
\parbox[t]{2mm}{\multirow{2}{*}{\scriptsize \rotatebox[origin=c]{90}{Phase II}}} 
& MECA-CTP & .471 & .569 & .536 & 55.00 & \multirow{2}{*}{+15} \\
& \method{} & \textbf{.499} & \textbf{.594} & \textbf{.562} & \textbf{59.41} & \\
\midrule
\parbox[t]{2mm}{\multirow{2}{*}{\scriptsize \rotatebox[origin=c]{90}{Phase III}}} & MECA-CTP & \textbf{.461} & .554 & .528 & \textbf{55.55} & \multirow{2}{*}{-3} \\
& \method{} & .447 & \textbf{.576} & \textbf{.529} & 54.34 &\\
\bottomrule
\end{tabular}
\label{comparison}
\end{table}

\noindent\textbf{New Diseases.} 
We select disease combinations $\mathcal{D}^{(j)}$ that appear 
ONLY in test sets but not in ANY training set, referring to them as ``New Diseases''. 
Detailed statistics are provided in Appendix~\ref{appd: new}. We compare our results with the previous SOTA, MEXA-CTP on the subset of new diseases. Table~\ref{comparison} shows, in addition to the previously used evaluation metrics, the gain in the number of correct predictions. \method{} consistently outperforms MEXA-CTP in F1, PR-AUC, and ROC-AUC in both Phase I and Phase II, especially achieving 13.63\% higher accuracy in Phase I and 8.02\% in Phase II, while matching MEXA-CTP’s performance in Phase III.

\subsection{Ablation Studies}
We conduct ablation studies for assessing (i) the importance of using multiple grouping blocks, (ii) the gains from the fine-tuning stage via LoRA, (iii) the necessity of incorporating a pre-training stage, and 
(iv) the proposed strategy for dataset usage in pre-training and fine-tuning stage. For these studies, we focus on the results for phase III, using the best hyperparameters obtained in Section~\ref{sec:settings}, unless stated otherwise.

\begin{table}[H]
\caption{Ablation studies for the Grouping Block configuration. $S$ indicates the number of the self-attention layer(s) in a grouping layer, and $G$ indicates the number of the grouping layer(s) in a grouping block. \method{} uses $G=3, S=2$.
}
\centering
\setlength\tabcolsep{5pt}
\begin{tabular}{ccccc}
\toprule
Method &  F1 & PR-AUC & ROC-AUC \\
\midrule
G=1, S=2& .797$\pm$.011 & .808$\pm$.015 & .611$\pm$.015\\
G=2, S=2 & .836$\pm$.009 & .803$\pm$.014 & .642$\pm$.011\\
\midrule
G=3, S=0 & .612$\pm$.011 & .669$\pm$.015 & .615$\pm$.023\\
G=3, S=1 & .773$\pm$.018 & .813$\pm$.015 & .618$\pm$.016\\
G=3, S=3 & .692$\pm$.015 & .857$\pm$.006 & .698$\pm$.015\\
\method{} & \textbf{.861} $\pm$.014& \textbf{.860}$\pm$.014 & \textbf{.702}$\pm$.018 \\
\bottomrule
\end{tabular}
\label{tab:ab}
\end{table}

\noindent\textbf{Grouping Block Setting.} We consider two main configurations for the setting of Grouping Block: (i) the number of $S$ self-attention layer(s) in a grouping layer, where $S \in \{0, 1, 2, 3\}$, and (ii) the number of $G$ grouping layer(s) in a grouping block, where $G \in \{1, 2, 3\}$. Our design, with $G=3, S=2$, yields the best performance compared to other configurations.
In all cases, the final number of the centroids is fixed to 25. For example, $G = 3, S=2$ indicates that, in each grouping layer, the cross-attention layer is followed by two self-attention layers, and for each grouping block, \method{} is initialized with 100 centroids, then halved to 50 centroids, and subsequently to 25 centroids.

\begin{table}[H]
\caption{Ablation studies for importance of introducing LoRA in fine-tuning stage. \method{} uses $\mathrm{head
}+\mathrm{X_{\mathrm{LoRA}}}+\mathrm{S_{\mathrm{LoRA}}}$.}
\centering
\setlength\tabcolsep{5pt}
\begin{tabular}{lccc}
\toprule
Method &  F1 & PR-AUC & ROC-AUC \\
\midrule
$\mathrm{head}$-$\mathrm{only}$ & .806 $\pm$.012& .857$\pm$.014 & .675$\pm$.011 \\
$\mathrm{head}$+$\mathrm{X_{\mathrm{LoRA}}}$ & .822 $\pm$.012& .856$\pm$.014 & 
.681$\pm$.024 \\
$\mathrm{head}$+$\mathrm{S_{\mathrm{LoRA}}}$ & .851 $\pm$.009& \textbf{.860}$\pm$.012 & .700$\pm$.016 \\
\method{} & \textbf{.861} $\pm$.014& \textbf{.860}$\pm$.011 & \textbf{.702}$\pm$.018 \\

\bottomrule
\end{tabular}
\label{tab:ab-ft}
\end{table}

\noindent\textbf{Fine-tuning with LoRA.} To further improve performance, we incorporate LoRA layers into the DM branch. We consider three alternative configurations for PEFT: (i) training only the parameters of the prediction head ($\mathrm{head}$-$\mathrm{only}$); (ii) training both the prediction head and LoRA layers in the cross-attention layer (head + $\mathrm{X_{\mathrm{LoRA}}}$); and (iii) training both the prediction head and LoRA layers in the self-attention layers for embedding enrichment (head + $\mathrm{S_{\mathrm{LoRA}}}$). 
As shown in Table~\ref{tab:ab-ft}, training the prediction head and both LoRA layers from the grouping layer (i.e., as done in \method{}) results in a significant performance boost, with a 1.2\% improvement in F1 score and a 2.9\% in ROC-AUC over the second best configuration.

\noindent\textbf{Training strategy.} We evaluate two alternative training strategies and compare them with our model: (i) using the pre-trained model directly without further fine-tuning and evaluating pair-matching results, and (ii) fine-tuning the model without a pre-training stage. As shown in Table~\ref{tab:ab-ts}, our training strategy (pre-training + fine-tuning) consistently outperforms the other two approaches.

\begin{table}[H]
\caption{Ablation studies for the necessity of combining both pre-training and fine-tuning stages.}
\centering
\setlength\tabcolsep{5pt}
\begin{tabular}{lccc}
\toprule
Method &  F1 & PR-AUC & ROC-AUC \\
\midrule
w/o fine-tuning & .357 $\pm$.008 & .466 $\pm$.009 & .514 $\pm$.010 \\
w/o pre-training & .800 $\pm$ .013 & .766 $\pm$ .12  & .680 $\pm$ .010 \\
\method{} & \textbf{.861} $\pm$.014& \textbf{.860}$\pm$.011 & \textbf{.702}$\pm$.018 \\
\bottomrule
\end{tabular}
\label{tab:ab-ts}
\end{table}

\begin{figure}[t]
\centering 
\includegraphics[width=\linewidth]{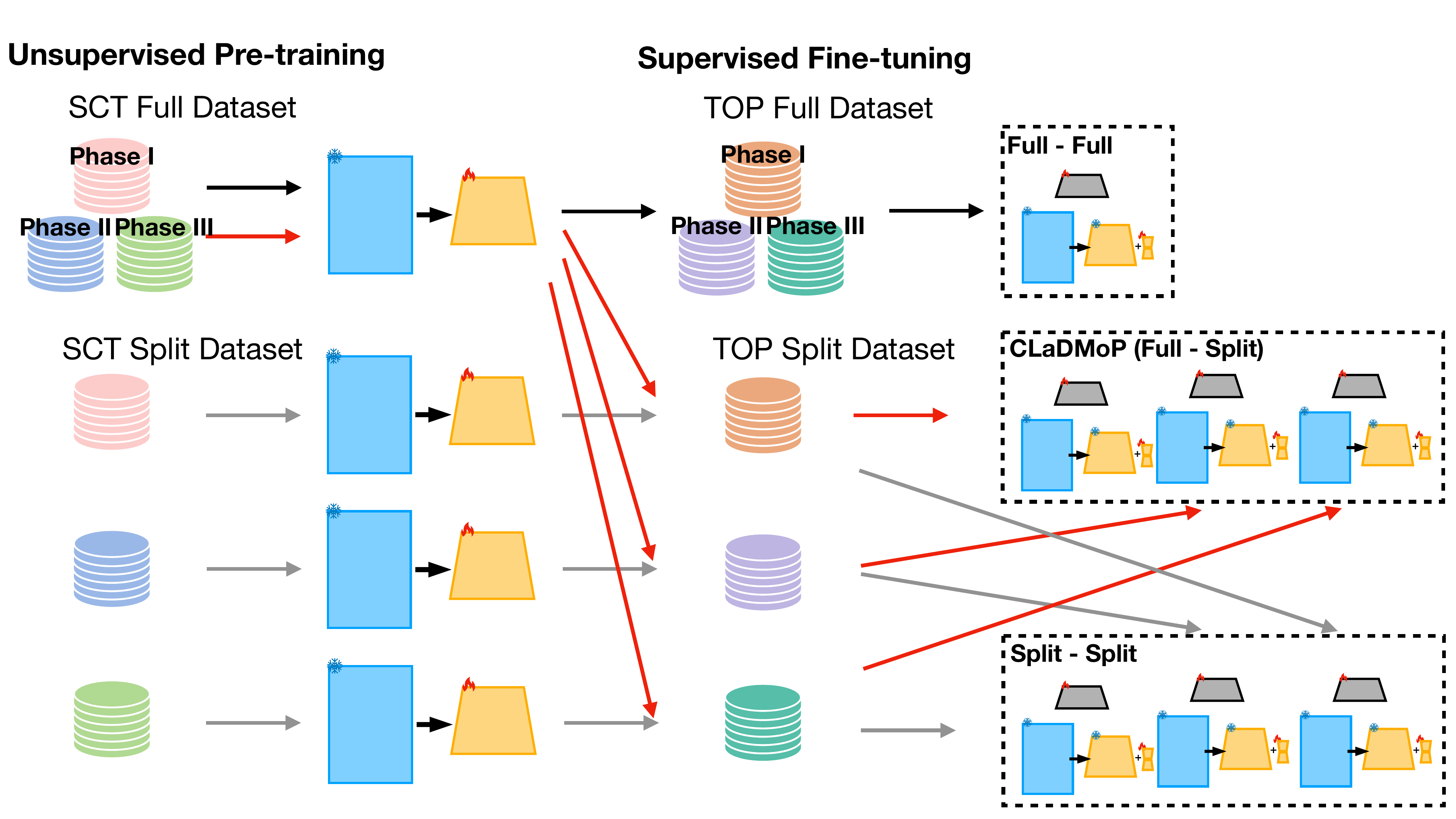}
\caption{We show three different training strategies for
model development, showing the dataflow during both pretraining and fine-tuning stages. The arrows represent the
path of the data usage for each strategy: (i) Split-Split Training Strategy, where separate models are trained on different
phases in SCT and then fine-tuned on the corresponding
phases in TOP; (ii) Black Arrows: Full-Full training strategy,
where the entire SCT is used for pre-training and entire TOP
is used for fine-tuning; (iii) Red Arrows: Full-Split Training Strategy, where the entire SCT dataset is used for pretraining, while fine-tuning is performed on phase-specific
TOP dataset.}
\label{datasplit}
\end{figure}

\noindent\textbf{Dataset Usage in Pre-Training and Fine-Tuning.} 
We analyze the impact of alternative strategies for using the \textsc{SCT} data during pre-training and the \textsc{TOP} data dataset during fine-tuning. In the pre-training stage, we explore two approaches: (i) \underline{split} the \textsc{SCT} data by trial phase to train phase-specific models, or (ii) use the \underline{full} data (i.e., including all samples across phases I, II and III) to train a single, general model. In the first case, it is natural to fine-tune each model variant by also splitting the TOP data accordingly--we call this strategy \textbf{Split-Split}. In the second case, we can take the phase-agnostic model and fine-tune it on the \textsc{TOP} full data--we call this strategy \textbf{Full-Full}, instead of splitting the TOP data as done in \method{}.
A sketch of the three strategies can be found in Fig.~\ref{datasplit}.

\begin{figure}[t]
\centering 
\includegraphics[width=\linewidth]{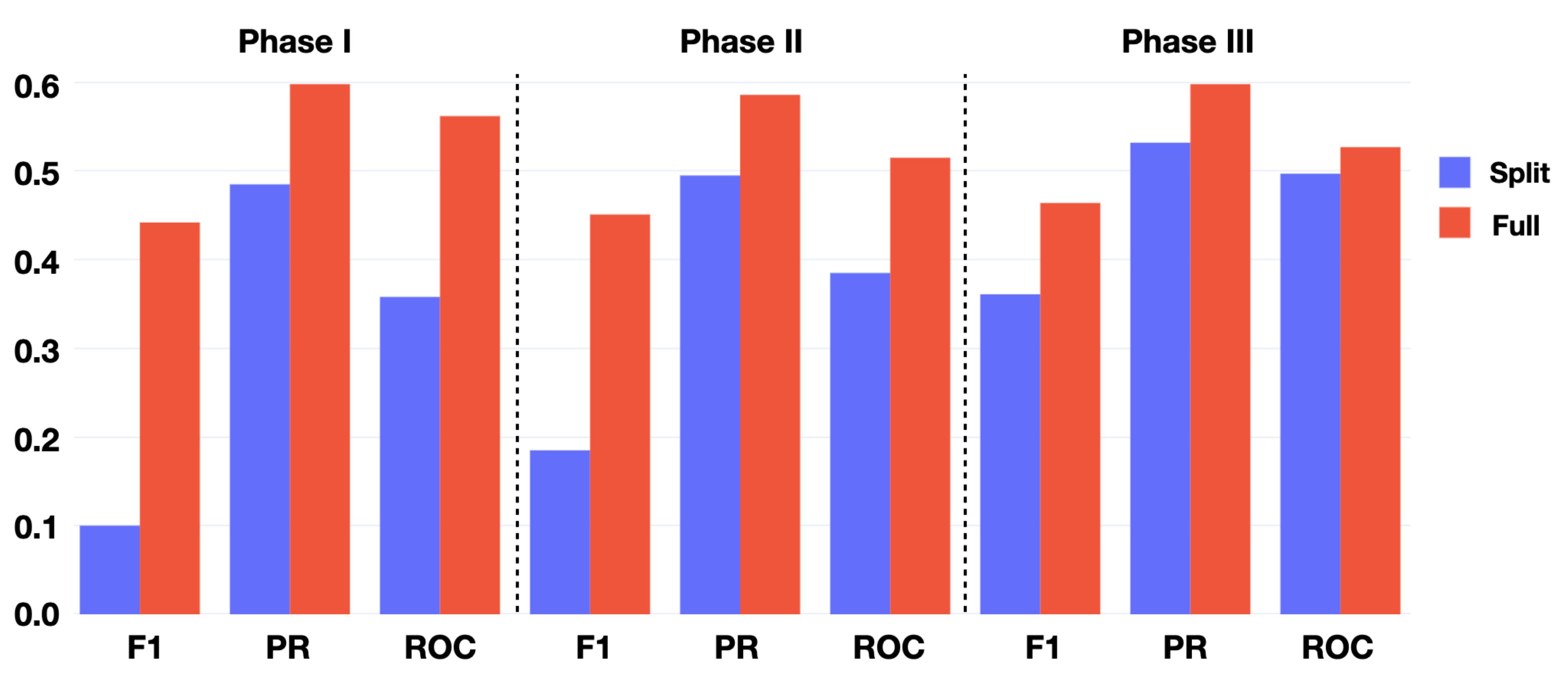}
\caption{Pre-trained models performance using SCT \textit{split} by clinical phase vs.\ the \textit{full} SCT dataset (which benefits from having more data).} 
\label{pre}
\end{figure}

\noindent\textit{Pre-training.} To highlight the advantages of using a large dataset for the pre-training stage, we compare models trained on \textsc{SCT} split dataset with the model trained on \textsc{SCT} full dataset, as shown in Fig.~\ref{pre}. \method{}'s strategy, i.e., training on \textsc{SCT} full, achieves improvements of 342\%, 144\%, 28.5\% in F1, 23.3\%, 18.4\%, 12.4\% in PR-AUC and 57.0\%, 33.8\%,  6.0\% in ROC-AUC for phases I, II, and III, respectively. The benefit of \textsc{SCT} full dataset is particularly significant when there are limited samples available in each phase.

\begin{figure}[t]
\centering 
\includegraphics[width=\linewidth]{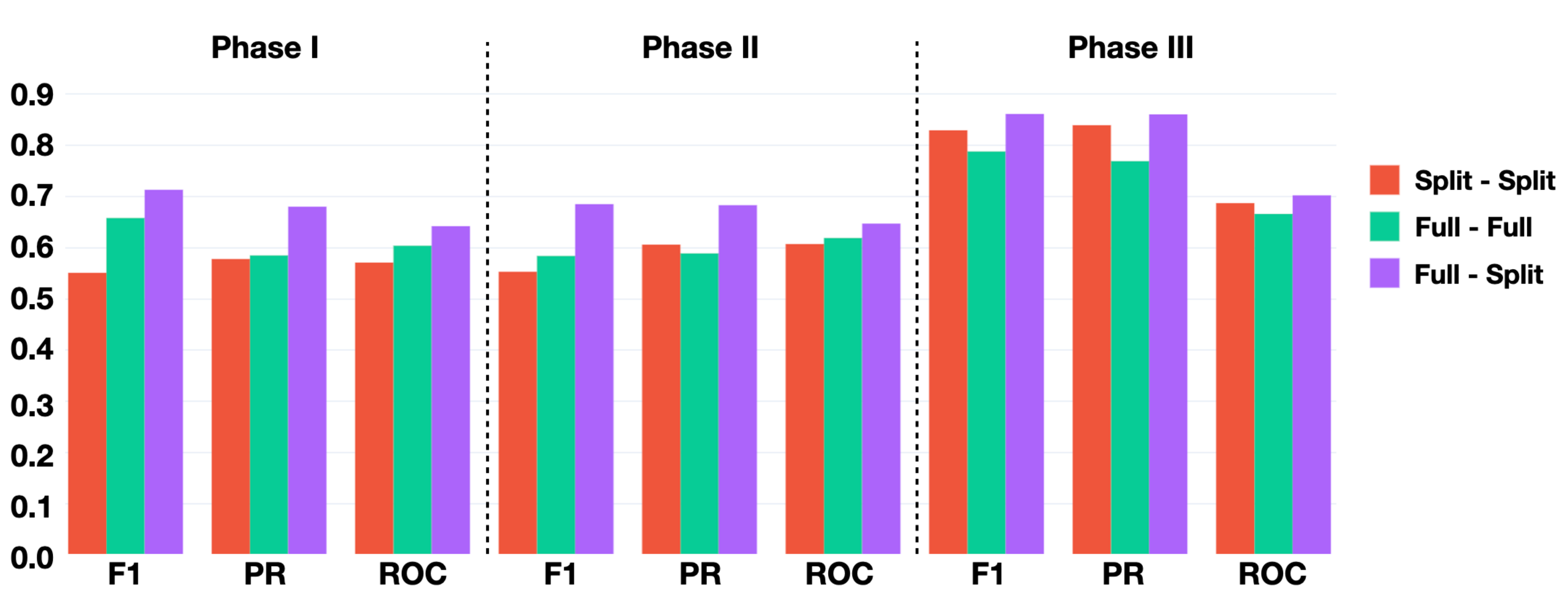}
\caption{(PEFT) fine-tuned models performance for three data usage strategies: 1$^\textrm{st}$ term in the key indicates if models were pre-trained on SCT split or SCT full; 2$^\textrm{nd}$ term indicates if they were fine-tuned on TOP split or TOP full. We found that Full - Split (\method{}'s strategy) performed best.}
\label{post}
\end{figure}

\noindent\textit{PEFT.} 
We compare fine-tuned models using alternative data usage strategies---namely, Full-Full and Split-Split--- with \method{}. Our training strategy (i.e., pre-training with the \textsc{SCT} full dataset and fine-tuning with the \textsc{TOP} split dataset) consistently achieves the best performance across phases I, II, and III, as shown in Fig.~\ref{post}.


\subsection{Complexity
Analysis} 
\label{sec:complexity}
As a transformer-based model, \method{}'s time complexity for training is $\mathcal{O}\left(d n^2 \right)$, where $d$ is the hidden size of the attention model, $n$ is the sequence length. 
Since the LLM branch uses a decoder-only model, in most cases the number of tokens from this branch largely exceeds those fused with the DM branch. Therefore,
\method{} fuses information from the LLM branch into the DM branch, 
causing the sequence length to grow linearly with the number of fusion steps. As a result, the attention computation in the DM branch scales quadratically.
However, our grouping blocks help maintain the token sequence length at a predefined value of 25, preventing quadratic growth in computation and significantly reducing the computational cost~\cite{zhang2021stabilized}.

\section{Related Work}
\noindent\textit{Clinical Trial Outcome Prediction.}
Many studies have focused on predicting individual patient outcomes within clinical trials through patient retrieval and enrollment information, rather than predicting overall trial success. For instance, Doctor2Vec~\cite{biswal2020doctor2vec} learns representations for medical providers and trials from electronic health records (EHR) and trial descriptions to address issues like recruitment in less populated countries. DeepEnroll~\cite{zhang2020deepenroll} encodes enrollment criteria and patient records into a shared latent space for matching, while COMPOSE~\cite{gao2020compose} uses medical ontology-based patient records and eligibility criteria embeddings for dynamic patient-trial matching. 
More recently, a few methods were proposed to directly predict clinical trial outcomes based on drug molecules, target diseases, and eligibility criteria. 
Hong et al.~\cite{hong2020predicting} focused on forecasting clinical drug toxicity using features related to drug and target properties, employing an ensemble classifier of weighted least squares support vector regression. RS-RNN~\cite{qi2019predicting} predicted phase III outcomes based on phase II results by considering time-invariant and time-variant variables. EBM-Net~\cite{jin2020predicting} inferred clinical trial outcomes by unstructured sentences from medical literature that implicitly contain PICOs (Population, Intervention, Comparison and outcome). More recently, HINT~\cite{fu2022hint} incorporated drug molecule features, target diseases, and eligibility criteria to design a neural network which consists of several modules connected as a graph encoding human priors.
MEXA-CTP~\cite{zhang2025mexa} predicts clinical trial outcome using same input as HINT but with minimal human efforts by learning to select a relevant subset of
tokens via mode experts. Our method, \method{}, incorporates a pre-training stage~\cite{hu2023self} using unlabeled data~\cite{zhang2021lancet,hu2023uce}, followed by a fine-tuning stage to further enhance performance. This two-stage training approach significantly improves the robustness of the embeddings and the model's predictive power. Meanwhile \method{} utilize the grouping layers to reduce computational costs~\cite{hu2024only}.

\noindent\textit{Large Language Models for Clinical Trials.} 
There is significant interest in leveraging Large Language Models (LLMs) for various prediction tasks in the context for clinical trials~\cite{frontiers2023}. They have been applied with mixed success for structuring eligibility criteria through logic operators and matching patients to clinical trials~\cite{mlhc2023,amia2023}, and are currently under consideration for assisting in clinical trial planning, coding-free text narratives and other side tasks. 
There is significant interest in leveraging Large Language Models (LLMs) for various prediction tasks in the context for clinical trials~\cite{frontiers2023}. They have been applied with mixed success for structuring eligibility criteria through logic operators and matching patients to clinical trials~\cite{mlhc2023,amia2023}, and are currently under consideration for assisting in clinical trial planning, coding free text narratives and other side tasks.
For clinical trial outcome prediction, however, it is not possible to directly use extant, subscription-based APIs (e.g., ChatGPT), due to their inability to handle some data modalities (e.g., drug molecules and ICD codes).
Nevertheless, recently released open-source LLMs (e.g., Llama~\cite{taori2023stanford}) can be, in theory, trained to deal with such inputs~\cite{naturebio2023,liu2025urbanmind,wu2024rose}. However, a number of formidable challenges must be overcome first: (i) the immense amount of data required to train large models from scratch, since models pre-trained on human language cannot be directly fine-tuned for this purpose; (ii) the risk of hallucinations, which require finding new ways of grounding the output produced by LLMs; (iii) the societal biases learned during training, which can lead to unfair outcomes, especially in critical areas like healthcare~\cite{salavati2024reducing}. As a result, there is still much debate around the regulation of the use of such models in the medical domain~\cite{lancet2024}.

\section{Conclusion}
Our approach demonstrates the potential of self-supervised learning to improve clinical trial outcome prediction with limited labeled data. With a two-branch architecture, we employ contrastive learning between LLM branch and DM brach using an adapted InfoNCE loss, allowing the model to learn robust task-agnostic representations. To support pre-training, we collect a new dataset, titled Successful Clinical Trials (\textsc{SCT}). Additionally, we leverage intermediate embeddings from the LLM branch through efficient information-fusion blocks (grouping blocks). Each grouping block consists of several grouping layers, which progressively reduce the token sequence length, enabling enhanced performance while minimizing computational costs. 
Our pre-trained model demonstrates significant improvements over established zero-shot and few-shot baselines based on evaluation across three clinical trial phases. We further improve the performance on the Trial Outcome Prediction (\textsc{TOP}) benchmark by training LoRA layers on the DM branch and a prediction head. \method{} achieves up to 10.5\% and 3.6\% improvements in PR-AUC and ROC-AUC, respectively, while maintaining a comparable F1 score to MEXA-CTP. We also analyze the accuracy and the absolute gain in correct predictions for new diseases compared to the previous state-of-the-art MEXA-CTP, showcasing that \method{} can learn generalizable embeddings via task-agnostic loss and easily adapt to new drug molecules. This method sets a promising direction for further advancements in clinical trial prediction using deep learning.

\newpage
\bibliographystyle{ACM-Reference-Format}
\bibliography{reference}

\newpage
\appendix
\section{Encoding Module.} 
\label{appd: EM}
\subsection{Drug Molecules Embedding.}
\label{appd: DM encoding}
We observe that although drug molecules may differ, they often share some SMILES segments. Therefore, we create molecule embeddings by intelligently combining their SMILES segment representations as obtained by DeepChem~\cite{ramsundar2018molecular}. To improve efficiency, we build an embedding dictionary $\textsc{Dict}_\mathrm{emb}$ for each SMILES segment,
\begin{equation}
    \textsc{Dict}_\mathrm{emb} = \{ e_{s_1}, e_{s_2}, \cdots, e_{s_{V}}\}, \quad  e_{s_{k}}= \textsc{DeepChem} (s_k)
\end{equation}
where $s_k; \, k \in [1..V]$ is a SMILES segment and $e_{s_k}$ is the corresponding representation pre-computed by $\textsc{DeepChem}$. Molecules embeddings are represented as a sequence of tokens from $\textsc{Dict}_\mathrm{emb}$, denoted by $U_{\mathcal{M}^{(j)}}$.

\subsection{Target Diseases Embedding.} 
\label{appd: TD encoding}
Each target disease in a trial is represented using the ICD-10 code tree, which expresses how different diseases and related within the disease classification system. We encode the structural information of each code using the $\textsc{icdcodex}$ package:
\begin{equation}
    \begin{split}
    U_{\mathcal{D}^{(j)}} =  
    \{e_{d_1^{(j)}}, e_{d_2^{(j)}}, \cdots, e_{d_{D_j}^{(j)}}\}, &\quad e_{d_{i}^{(j)}} = \textsc{icdcodex} (d_i^{(j)}), \\ i &\in [1..D_j],
    \end{split}
\end{equation}
where $U_{\mathcal{D}^{(j)}}$ represents a sequence of tokens for target disease embeddings.

\subsection{Information Enrichment}
We build a light weighted model for the Disease-Molecule branch. We enrich the representation from molecules embedding and disease embedding separately
 via self attention, 
\begin{equation}
U_{\mathrm{enriched}}^{\mathrm{domain}} = \mathrm{softmax}\left(\frac{U^{\mathrm{domain}}W^Q (U^{\mathrm{domain}}W^K)^\top}{\sqrt{d_k}}\right) U^{\mathrm{domain}}W^V,
\end{equation}

where $\mathrm{domain} \in \{\mathcal{M}, \mathcal{D}\}$.

\section{Dataset.}
\label{appd: Dataset}
\subsection{TOP dataset Statistics.}
\label{appd: top}
We provide the statistics for the \textsc{TOP} benchmark dataset in the table below.
\begin{table}[H]
\caption{\textsc{TOP} benchmark dataset statistics.}
\label{statistics}
\centering
\setlength\tabcolsep{15pt}
\begin{tabular}{ccccl}
\toprule
Phase &  Training & Test & Success Ratio\\
\midrule
I & 1,164 & 627 & 68\% \\
II & 4,451 & 1654 & 35\% \\
III & 4,313 & 1,146 & 30\% \\
\bottomrule
\end{tabular}
\end{table}

\subsection{New Diseases Statistics.}
\label{appd: new}
New diseases are defined as combinations of diseases present in the test set but absent from all phases of the training set.
\begin{table}[H]
\caption{New Diseases statistics.}
\label{statistics_new}
\centering
\setlength\tabcolsep{15pt}
\begin{tabular}{ccccl}
\toprule
Phase &  New Disease & Test & Percentage \\
\midrule
I & 138 & 627 & 22\% \\
II & 340 & 1654 & 21\% \\
III & 198 & 1,146 & 17\% \\
\bottomrule
\end{tabular}
\end{table}

\section{Baselines.}
\subsection{Pre-training baselines.}
\label{app: pre-training baselines}
\textbf{Model description}
\begin{itemize}
\item \textbf{ANIL} focuses primarily on updating the last layer of the model (i.e., the task-specific part) while leaving the rest of the model's parameters unchanged.
\item \textbf{MAML} is to train a model on a distribution of tasks, such that it can adapt quickly to new, unseen tasks with minimal gradient updates.
\item \textbf{BioGPT} is a question answering model pre-trained on large-scale biomedical corpora, including research papers, clinical trial reports, medical textbooks, and other domain-specific text sources.
\end{itemize}
\noindent\textbf{ANIL \& MAML.}
\noindent \textit{Implementation.} We follow the approach outlined in the ANIL paper~\cite{raghu2019rapid} and implement the code using PyTorch. All models consist of a 3-layer feed-forward neural network with ReLU activation functions.

\noindent \textit{Optimization.} We applied class weight parameters to recalibrate the loss when the respective functions provided this hyperparameter.
To gain best hyperparameters for each function in each phase, we
employed cross-validation while training.

\noindent\textbf{BioGPT}. The conversation template with BioGPT model. 

\begin{algorithm}[H]
    \caption{Conversation with LLMs}
    \begin{algorithmic}[1]
        \STATE \textbf{System:} You are a clinical researcher.
        \STATE \textbf{System:} I will give you CRITERIA, DISEASE, and DRUG for a clinical trial. Please give me an answer with ``YES'' or ``NO''directly.
        \STATE \textbf{User:} ELIGIBILITY: [eligibility criteria]
        \STATE \textbf{User:} DISEASE: [name of target diseases]
        \STATE \textbf{User:} DRUG: [name of drug compounds]
        \STATE \textbf{Prompt:} Does the combination of the CRITERIA, DISEASE, and DRUG represent a good match in [phase] for predicting clinical outcomes?
    \end{algorithmic}
\end{algorithm}

\subsection{Fine-tuning Baselines.} \label{app: fine-tuning baselines}
\textbf{ML-based methods.}
\noindent \textit{Implementation.} We utilized scikit-learn packages for all machine learning baselines, including Logistic Regression, Random Forest, k-Nearest Neighbor + Random Forest, XGBoost, Adaptive Boosting, a 3-layer Feed-Forward Neural Network.


\noindent \textit{Data Preprocessing.}  We utilize the same encoding module outlined in MEXA-CTP~\cite{zhang2025mexa}. For handling missing values, if the method involves k-nearest neighbors, we will generate k clusters using the non-missing values and predict the missing value based on the centroid of the corresponding cluster. For methods that do not incorporate clustering, we will substitute missing values with zeros. Additionally, we will pad and chunk the array to ensure same size, using the normalized array as input for all the baseline models.

\noindent \textit{Optimization.} We applied class weight parameters to recalibrate the loss when the respective functions provided this hyperparameter. To gain best hyperparameters for each function in each phase, we employed 5-fold cross-validation while training.

\noindent \textbf{HINT.} We strictly follow the HINT paper and their official GitHub\footnote{\url{https://github.com/futianfan/clinical-trial-outcome-prediction}}.

\noindent \textbf{MEXA-CTP}. We strictly follow the MEXA-CTP paper and their official GitHub\footnote{\url{https://github.com/murai-lab/}}.
\subsection{Batch Size for Pre-training}
\label{pre-bs}
The influence of the batch size affect the pre-training performance. 

\begin{figure}[h]
\centering 
\includegraphics[width=\linewidth]{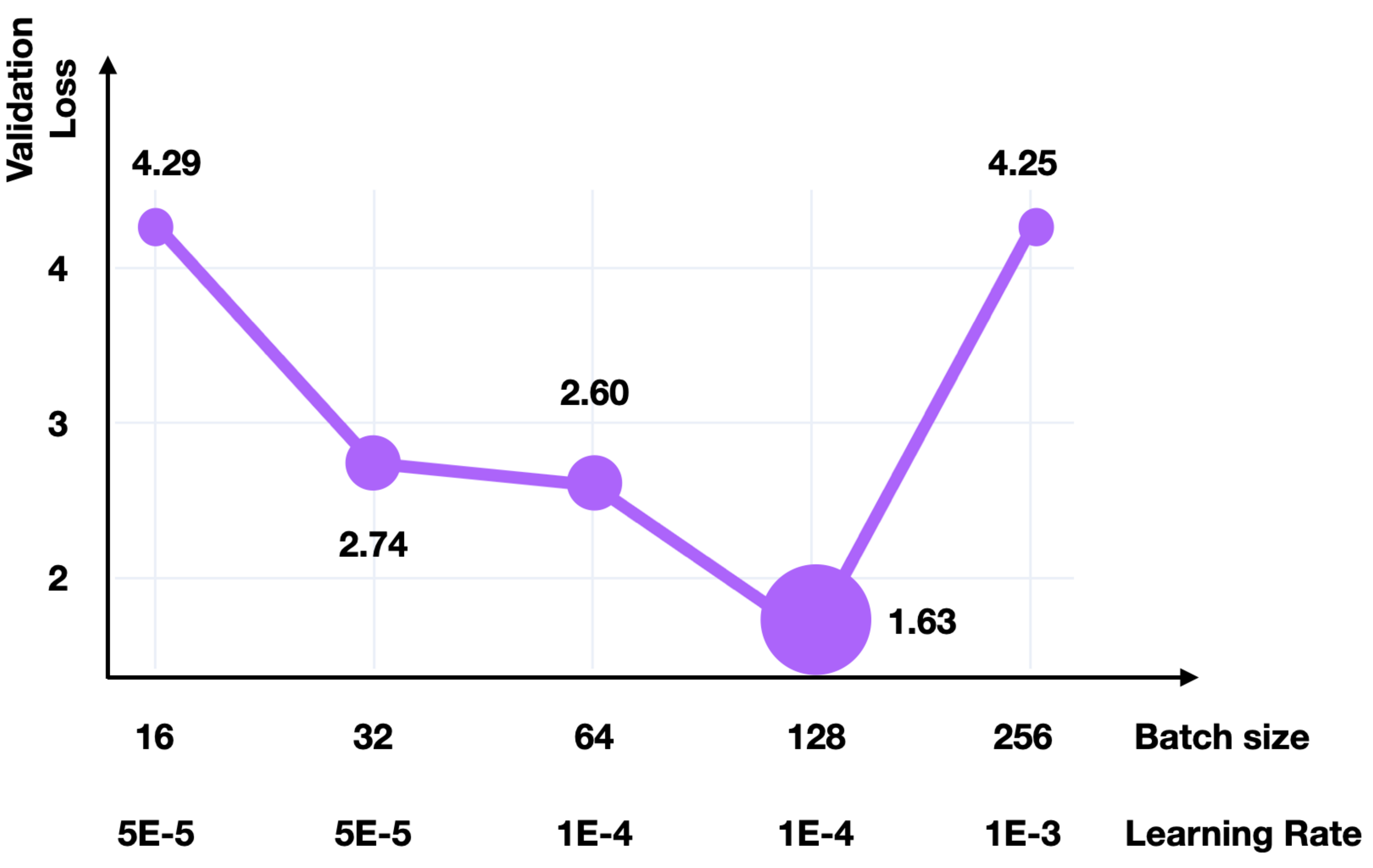}
\caption{Batch size vs. validation loss. We show the corresponding learning rate for each batch size at the bottom of the figure. The size of the ball indicates the number of epochs required to reach the minimum validation loss.} 
\label{ab}
\end{figure}

\noindent\textbf{Batch Size in Pre-training.} In previous studies~\cite{oord2018representation, gao2021scaling, chen2022we}, batch size has been shown to be a crucial hyperparameter in contrastive-based pre-training. ave demonstrated that batch size is a critical hyperparameter in contrastive-based pre-training. To achieve high-quality pre-training results, we experimented with different batch sizes to assess their impact on performance. As shown in Fig.~\ref{ab}, we choose a batch size of 128, which required 167 epochs to converge to the minimum validation loss.

\end{document}